\def\year{2020}\relax
\documentclass[letterpaper]{article} 
\usepackage{aaai20}  
\usepackage{times}  
\usepackage{helvet} 
\usepackage{courier}  
\usepackage[hyphens]{url}  
\usepackage{graphicx} 
\usepackage{graphicx}  
\usepackage{amsmath}
\usepackage{amsfonts}
\usepackage[utf8]{inputenc}
\usepackage{amsmath,amssymb}
\usepackage{array}
\usepackage{graphicx}
\usepackage{multirow}
\usepackage{subfig}
\usepackage{todonotes}
\usepackage{listings}
\usepackage{makecell}
\title{Visual Story Generation Based on Emotion and Keywords}
\author{\Large \textbf{Yuetian Chen, Ruohua Li, Bowen Shi, Peiru Liu, and Mei Si}\\ 
Rensselaer Polytechnic Institute\\ 
110 8th street, 
Troy, New York 12180\\
\{cheny63, lir11, shib5, liup5, sim\}@rpi.edu 
}
 
\begin{document}

\maketitle

\def\mybar#1{
  {\color{brown}\rule{#1cm}{8pt}} #1}
  
\begin{abstract}
Automated visual story generation aims to produce stories with corresponding illustrations that exhibit coherence, progression, and adherence to characters' emotional development. This work proposes a story generation pipeline to co-create visual stories with the users. The pipeline allows the user to control events and emotions on the generated content. The pipeline includes two parts: narrative and image generation. For narrative generation, the system generates the next sentence using user-specified keywords and emotion labels. For image generation, diffusion models are used to create a visually appealing image corresponding to each generated sentence. Further, object recognition is applied to the generated images to allow objects in these images to be mentioned in future story development. 
\end{abstract}

\section{Introduction}
Storytelling has been an important way of communication, entertainment, and even making sense of the world. Stories can be told with just text or using the visual median. The use of visualizations can make stories more expressive and engaging. In this work, we propose a visual story co-creation system that leverages the power of large language models and diffusion models to create short visual stories with a user. \footnote{GitHub repository: \textit{https://github.com/Stry233/Visual-Story-Generation-Based-on-Emotional-and-Keyword-Scheme}.}

The example short stories we have in mind are from the ROCStories, a popular dataset for commonsense reasoning and narrative understanding~\cite{mostafazadeh2016corpus} named Story Cloze Test. Each story is composed of five sentences and follows a character through a series of events to an ending event or situation. When the data were collected, the crowdsource workers were instructed that ``the story should read like a coherent story, with a specific beginning and ending, where something happens in between'' \cite{mostafazadeh2016corpus}. These instructions make them good examples of meaningful short stories. While the stories in the dataset only have text, our proposed system aims to help people create similar stories with visualization. 

\begin{figure*}[h]
    \centering
    \includegraphics[width = 0.9\linewidth]{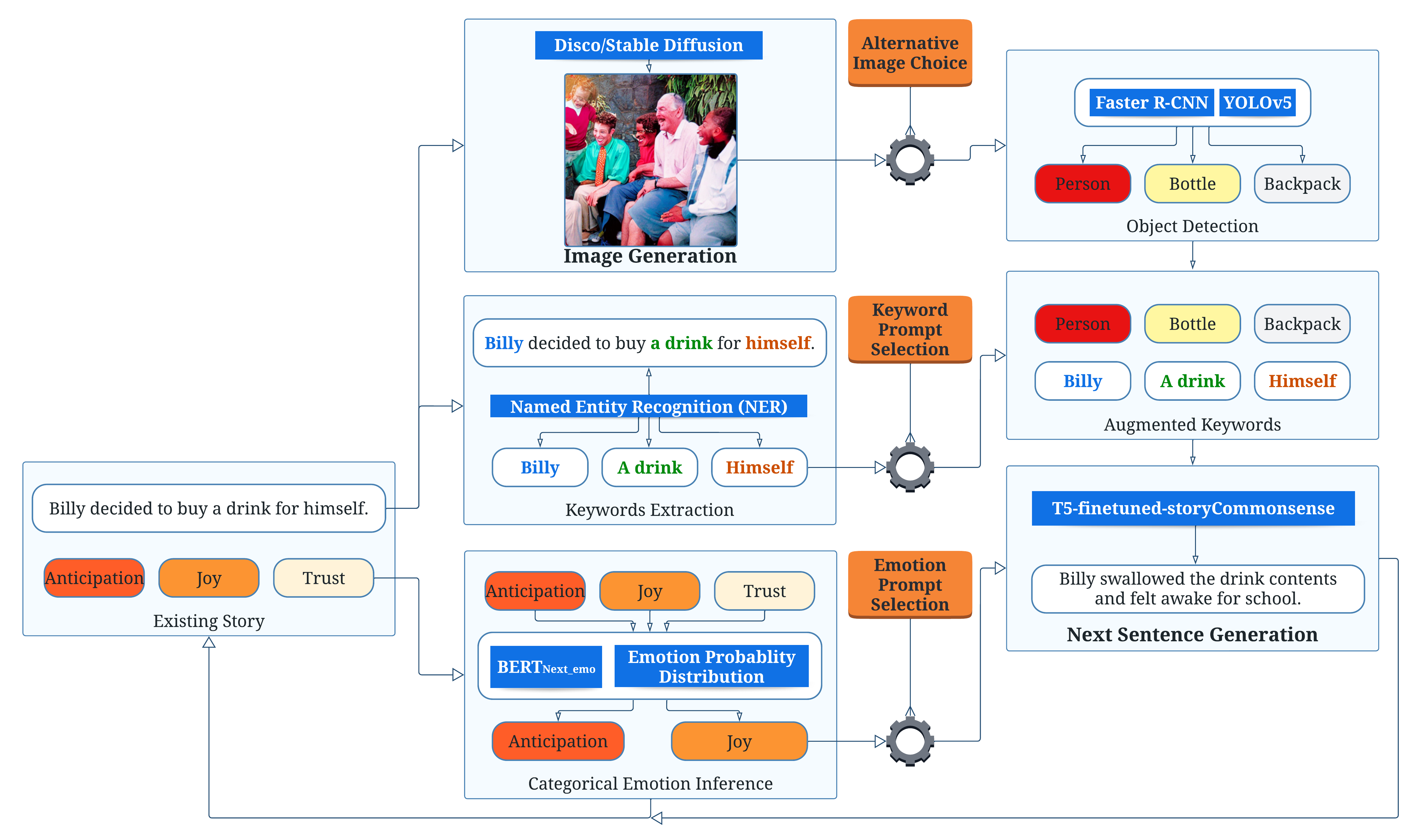}
    \caption{Generation Pipeline. 
    }
    \label{PipelineMain}
\end{figure*}

To co-create visual stories, the proposed system aims to provide plausible suggestions for story development while allowing users to override using their own suggestions. Figure~\ref{PipelineMain} shows the overall pipeline of the system. The system is iterative in nature. Based on the existing partial story, the system can suggest the characters' emotions and keywords included in the next sentence, such as an object, a location, etc. The user can accept them as defaults or provide their own. The system will then generate the sentence below based on the input parameters and the previously generated story. The user will then be presented with several visualizations for the new sentence and must choose one to proceed. Finally, the generated images will be subjected to object recognition so that the system can extract additional keywords and recommend their use in future storytelling.

\section{Related Work}
Automatic story generation has long been a challenging task in natural language processing, and such efforts date back to the 1970s~\cite{Meehan1977TALESPINAI}. Earlier attempts, such as symbolic planning systems~\cite{Riedl_2010}, utilize planning techniques to create plausible and coherent plots for stories. In addition, case-based or analogical modeling systems generate new narratives based on adapted existing stories~\cite{Gervs2005StoryPG}. However, while these approaches can create stories with impressive coherence and consistency, they often require extensive knowledge engineering and restrict to a limited domain. To address the issue, large crowd-sourced corpora of stories are used to help generate stories~\cite{10.1145/2362394.2362398,Li}. Modern approaches based on neural language models have been explored to generate plot-driven stories~\cite{fan-etal-2018-hierarchical}.
Moreover,~\cite{yang-etal-2019-enhancing-topic} incorporates commonsense knowledge into the generator using a novel memory mechanism and utilizes adversarial training to improve the diversity and originality of essay generation.~\cite{Tambwekar2019ControllableNS} leverages reward strategies during the generation to guide the language model toward generating a coherent story.~\cite{Liu_Li_Yu_Huang_Liu_Zhao_Yan_2020} proposes a character-centric story-generation model that can produce stories consistent with characters' profiles. 

In contrast to previous work, we propose an interactive visual story generation pipeline in which the user can specify keywords and emotions that will appear in the next sentence. The system generates the sentence and associated images for the user based on this information.

\section{Visual Story Co-creation Pipeline}
This section describes our approach to story generation pipelines and their interactive workflow, as shown in Figure~\ref{PipelineMain}. The pipeline is divided into two sections: next-sentence generation and image generation. The next sentence generation is in charge of producing a plausible story based on specified keywords and emotions, whereas image generation is in charge of producing visualization for the generated story. We can mine the generated images for additional keywords to include in future story development when it operates interactively.

\section{Next Sentence Generation}
To begin the co-creation process, the pipeline must be fed with the first sentence of the story. It will then suggest emotions for the characters as well as keywords for the next sentence, inviting the user to edit them. This process is repeated iteratively by the pipeline to build the entire story.

We trained a next-sentence emotion prediction model using the Story Commonsense dataset~\cite{commonsense}. Here, the existing story text was used as input, while the emotion labels from the next sentence were used as labels. We fine-tuned a BERT$_{\textsc{BASE}}$ model~\cite{devlin2018bert}, hereinafter noted as BERT$_{\textsc{Next Emo}}$, for this task. We provide the objects recognized from the generated images as the default for keyword suggestions. 

To enable the system to generate a sentence with the suggested keywords and emotions, we fine-tuned a pre-trained T5$_{\textsc{BASE}}$~\cite{raffel2020exploring} model using the ROCStories dataset\cite{mostafazadeh2016corpus}. The T5 model was pre-trained using the CommonGen dataset~\cite{lin2019commongen}, which enabled it to generate a sentence with a set of given keywords(hereinafter denoted as T5$_{\textsc{BASE Fine-tuned CommonGen}}$). Fine-tuning it with the ROCStories dataset enabled the T5 model to further learn about the story writing styles in ROCStories. 

\subsection{Keywords Extraction}
For training the model and evaluation, we extract keywords from the original five-sentence stories. We used the \texttt{SceneGraphParser()} from~\cite{wu2019unified} to parse sentences (in natural language) into scene graphs. The entities in the scene graphs become the keywords. For example, for the following sentence: 
\begin{center}
    \texttt{
    $\textsc{S}_1$: I brought the movie home and watched the whole thing.
    }
\end{center}
We are expected to generate the following result:
\begin{center}
    \texttt{'I, the movie, the whole thing'}
\end{center}

\subsection{Categorical Emotion Inference}
\begin{figure}
    \centering
    \includegraphics[width = 0.5\linewidth]{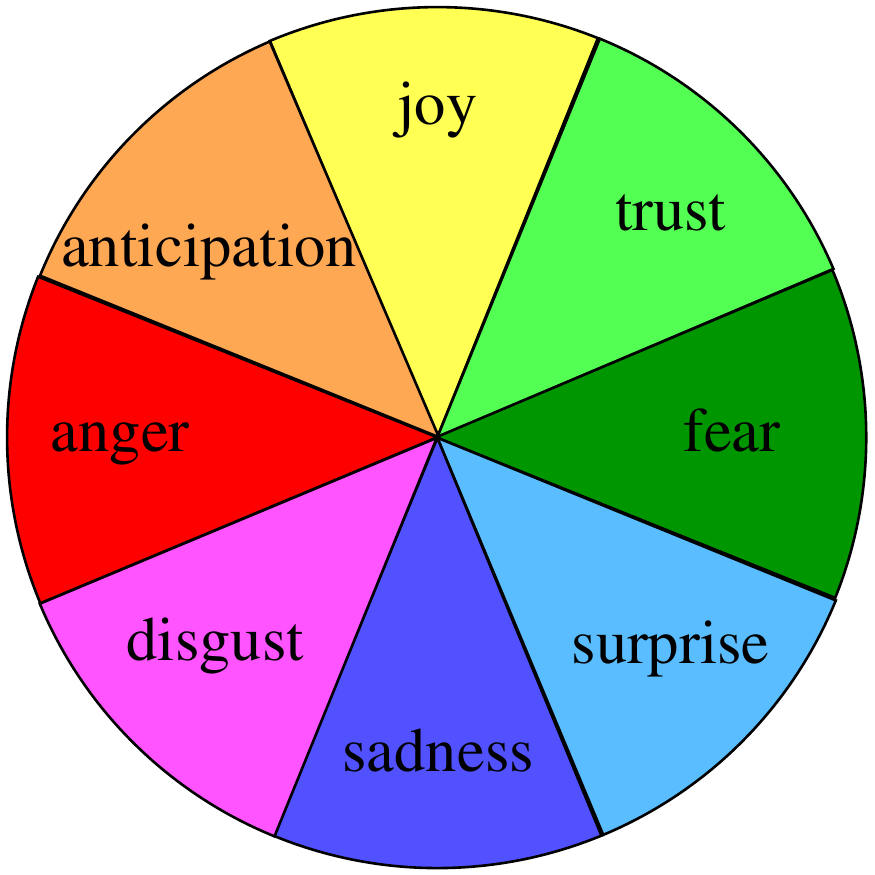}
    \caption{Plutchik basic emotions}
    \label{fig:plutchikLabel}
\end{figure}

We use Plutchik's Wheel of Emotions as our basic model of characters' emotions. The wheel includes eight emotions in pairs: joy/sadness, trust/disgust, fear/anger, and surprise/anticipation~\cite{plutchik1980general} as shown in Figure~\ref{fig:plutchikLabel}. We define emotion entries in a sentence as a real-valued, low-dimensional vector $\vec{\mathbf{C}}$.

\begin{align}
    \forall \vec{C} \in \mathbf{D}, \vec{C} &= \begin{bmatrix}
                                                e_{1} \\
                                                e_{2} \\
                                                \vdots \\
                                                e_{8}
                                                \end{bmatrix}, 
    e \in [0, 1]
\end{align}
\noindent Given the context of a story, i.e., all previous sentences $\vec{X} = \{x_1, x_2, x_3, \dots, x_m\}$, the goal is to predict the emotions $\vec{C_i}$ that will appear in the next sentence.


For example, for the following input sequence of sentence $\textsc{S}_1$ that serves as context, the model may predict the emotions in the next sentence to be \texttt{joy, anticipation, }and \texttt{trust}.

\begin{center}
    \texttt{
    $\textsc{S}_1$: He was hoping this year to be tall enough for the coaster.
    }
\end{center}

For preparing the training data, we obtained 17,910 pairs of story context and next-sentence emotions from the Story Commonsense dataset. The dataset was divided into training and validation sets in the ratio of 6:4. We then fine-tuned the BERT$_{\textsc{Next Emo}}$ model 
using the task of multi-label classification. The story context is the input, and the next-sentence emotions are the output. The maximum input length of the tokenizer was set to 120. The batch size was set to 16, and the learning rate was set to $6e-6$. We ran 16 epochs for training, which took about 45 minutes on a Tesla P100-PCIE-16GB GPU. The model achieved a Macro ROC-AUC score of 0.69 on prediction. The prediction results are not perfect. However, they are only used as suggestions, and the user can always overwrite them. 

\subsection{Next Sentence Generation with Prompts}
\begin{figure*}[h]
    \centering
    \includegraphics[width = \linewidth]{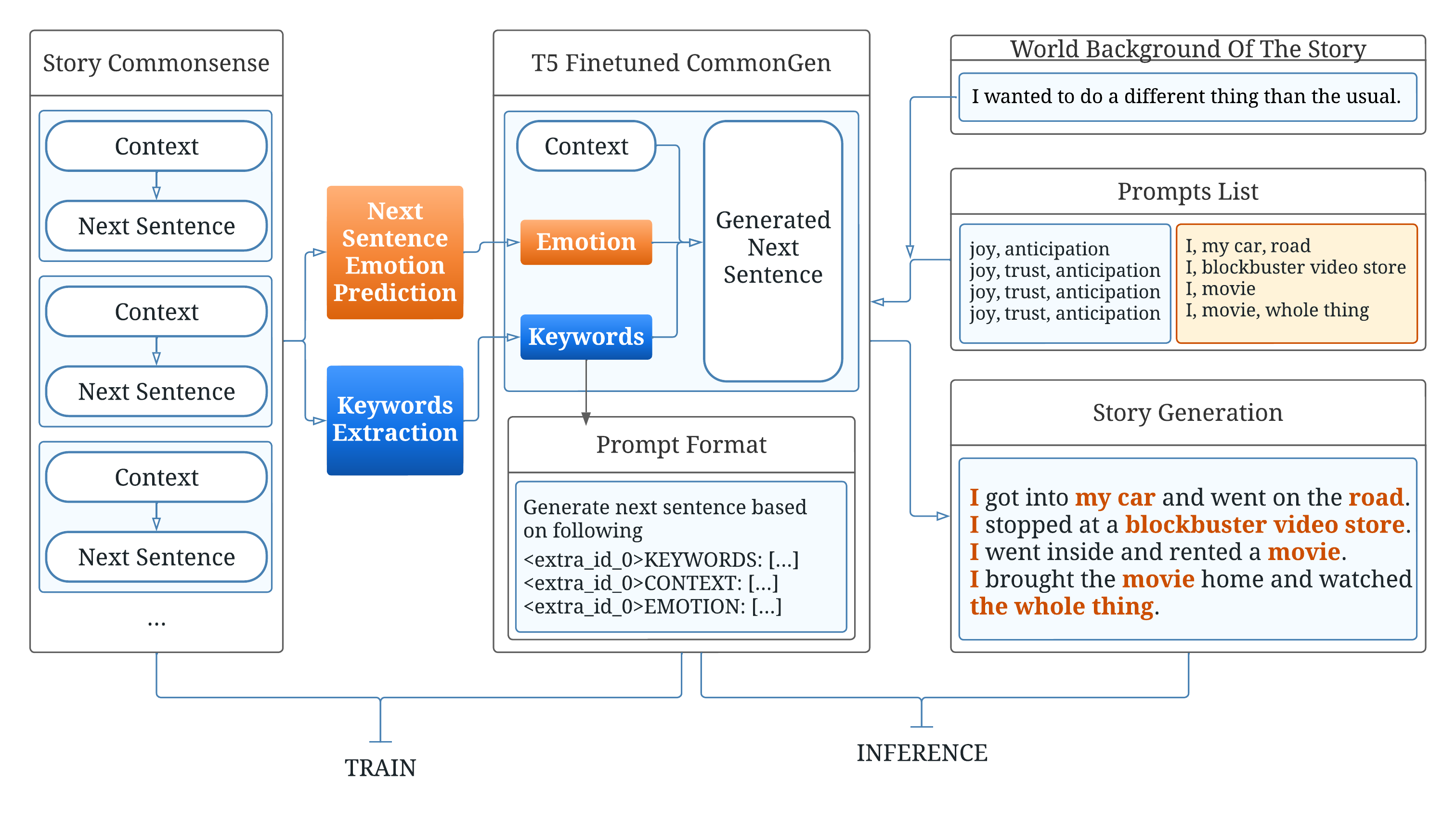}
    \caption{Training and inference pipeline for Next Sentence Generation.}
    \label{T5-TrainPipe}
\end{figure*}
An overview of our story generation model is given in Figure~\ref{T5-TrainPipe}. We constructed our training data using the stories from both the ROCStories and Story Commonsense datasets (denoted as $\mathbf{S_{total}}$ hereinafter). 

For a sample story from ROCStories -- $\vec{S} \in \mathbf{S_{total}}$, we define it as a 5-dimensional vector $\vec{S} = \{s_1, \dots, s_5\}$ where each $s_i$ represents the corresponding sentence in the story. 
%

For training our system to generate the next sentence, $s_{j+1}$ is fed into the emotion classifier as well as the keyword extractor for identifying the emotions and keywords in the next sentence. Based on story context and the additional knowledge input, T5$_{\textsc{BASE Fine-tuned CommonGen}}$ is trained on a triad of data: context, keyword, and emotion labels. 

In addition, we used the following prompting structure in both the training as well as inference processes:
\begin{lstlisting}
Generate next sentence based on following 
<extra_id_0>KEYWORDS: [...]
<extra_id_0>CONTEXT: [...]
<extra_id_0>EMOTION: [...]
\end{lstlisting}

We adopt T5$_{\textsc{BASE Fine-tuned CommonGen}}$ as the base language model for the training process. 
The CommonGen dataset contains pairs of sentences and keywords. The task is to generate a meaningful sentence that contains all the given keywords. 
The T5$_{\textsc{BASE Fine-tuned CommonGen}}$ is a T5 model fine-tuned using this dataset. For example, given the following input:
\begin{center}
    \texttt{
    X$_1$: <'Mike', 'tree', 'ground', 'hole'>
    }
\end{center}
$T5_{\textsc{Fine-tuned CommonGen}}$ may generate:
\begin{center}
    \texttt{
    Y$_1$: $\underset{\textsc{x}_0}{\text{[Mike]}}$ is digging a $\underset{\textsc{x}_3}{\text{[hole]}}$ in the $\underset{\textsc{x}_2}{\text{[ground]}}$ near a $\underset{\textsc{x}_1}{\text{[tree]}}$.
    }
\end{center}

By fine-tuning T5$_{\textsc{BASE Fine-tuned CommonGen}}$ based on the emotions and the keyword information as prompts, the model can more effectively capture hidden information in the story and thus improve the accuracy and variety of sentences. Compared to the baseline model fine-tuned only with context information, we found that, due to the lack of a priori knowledge of the link between emotions and events, the baseline model is highly susceptible to confounding empathy problems arising from different behavioral motivations. As shown in Table~\ref{storyExampleII}, the baseline model could not use the context effectively and correctly without the prompts. It generated repeated sentences. Further, it is unclear why the character -- Mary -- would feel happy about her situation. 

\begin{table}[h]
    \centering
    \begin{tabular}{p{0.45\linewidth}|p{0.45\linewidth}}
    \textbf{generated - prompt}                & \textbf{generated - no prompt}                \\ \hline
    Mary had been feeling depressed lately.                        & Mary had been feeling depressed lately. \\ \hline
    she decided to go see a psychiatrist.                          & She decided to go to a psychiatrist.    \\ \hline
    Psyched, her psychiatrist diagnosed her with depression and sent her to see. & She was diagnosed with schizophrenia. \\ \hline
    Medicant took her to get an antidepressant and prescribed her. & She was very happy.                     \\ \hline
    Thankfully it eventually made her feel better again.           & She was very happy.                    
    \end{tabular}%
    \caption{An example comparing the baseline model and model with emotion and keywords prompting}
    \label{storyExampleII}
\end{table}

In terms of implementation, we used the previous format to create a dataset with 18,680 entries. In an 8:2 ratio, we divided it into a training set and a validation set. The maximum lengths of the source and output text are set to 512 and 50, respectively. To control the diversity of the output content, we use top-K search in the decoding process, with $k=3$, and set the repetition penalty and length penalty to 2.6 and 1.0, respectively, for the same reason. In the training process, we set the batch size and learning rate to 10 and 5e-4, respectively, and introduced the early stop mechanism to prevent overfitting. We ran 16 epochs, and the training phase of the language model took about 10 hours on a Tesla P100-PCIE-16GB platform.

\section{Image Generation}
When image generation is added to text-based stories, it increases flexibility and creates a sense of immersion in the generated story plot. As a result, in addition to the text pipeline, we implement an image pipeline that generates additional keywords based on images as suggestions, potentially pushing the story forward in different directions. We separate the pipeline into two parts: image generation and object detection. We experimented with both Stable Diffusion~\cite{rombach2022high} and Disco Diffusion as our primary method for image generation, as they can produce fairly relevant figures with artistic quality. For object detection, we chose YOLOv5~\cite{glenn_jocher_2022_6222936} and Faster R-CNN~\cite{https://doi.org/10.48550/arxiv.1506.01497} as our models. The generated keywords will serve as suggestions for users to select from, as well as hints to guide the subsequent generation process.

\subsection{Image Generation}
To create images for the generated sentences, we utilize Disco Diffusion, a tool that combines both Diffusion, a mathematical process for removing noise from an image, and CLIP~\cite{https://doi.org/10.48550/arxiv.2103.00020}, a model for labeling images. Image generation is an iterative process in Disco Diffusion, in which CLIP evaluates the intermediate image based on the text prompt and provides a guideline for Diffusion to process at each step.

\begin{figure}%
    \centering
    \subfloat[\centering Prompt without adding ``country view" ]{{\includegraphics[width=3.6cm]{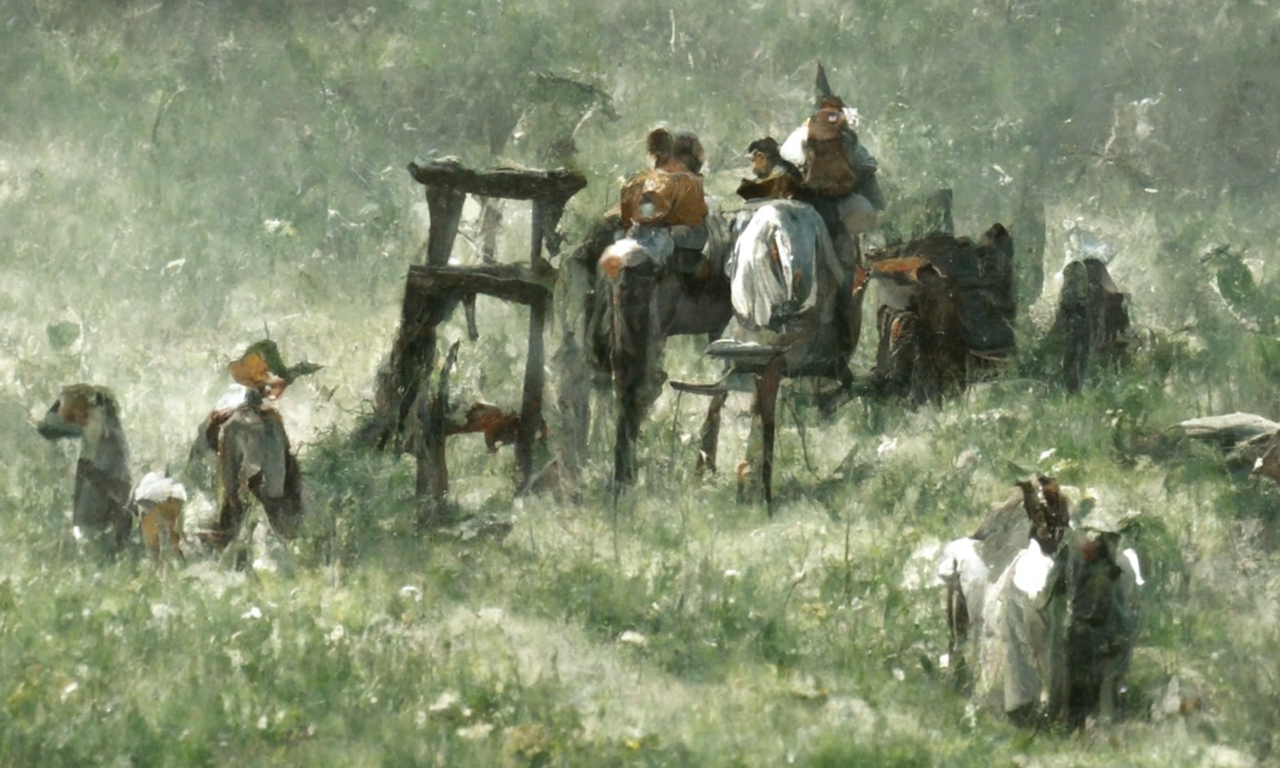} }}%
    \qquad
    \subfloat[\centering  Prompt adding ``country view"]{{\includegraphics[width=3.6cm]{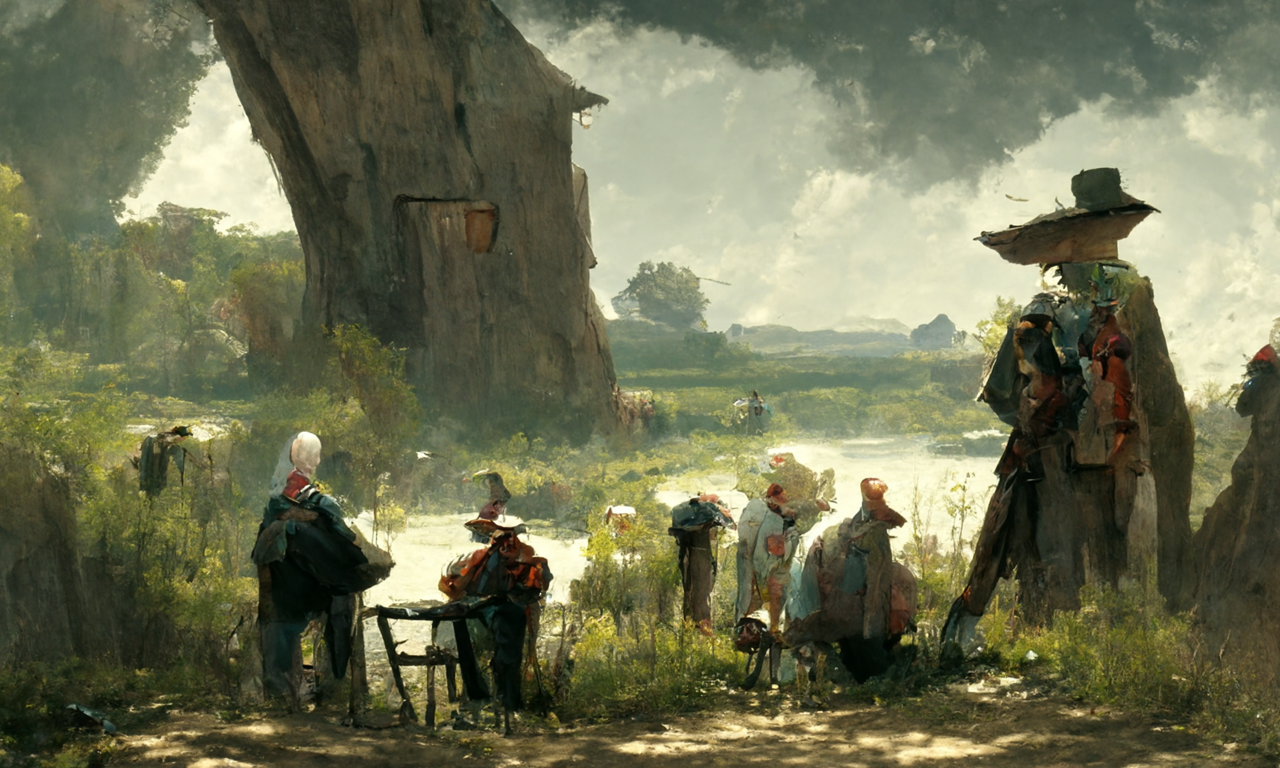} }}%
    \caption{Prompt: ``Ray gathered his friends to tell them a funny joke he heard," generated using Disco Diffusion}%
    \label{countryview}%
\end{figure}

Disco Diffusion has been extensively researched as a means of generating creative art. In practice, however, we found it difficult to create a good image based solely on one line of a story. Although Disco Diffusion has a portrait generator, the model finds it difficult to generate images that describe actions in the format of "someone is doing something." To address this, we include user-specific keywords in the text prompt, including the artist's name and background information. For example, we added "Carl Spitzweg" as the artist and "country view" as the background information in our example. Disco Diffusion can generate images that are consistent with the artist's artistic styles, such as shadow, color, and strokes, by including the artist's name in the text prompt. Furthermore, we fill in background information that can add more relevant elements to the generated image. As shown in Figure~\ref{countryview}, the prompt "Ray gathered his friends to tell them a funny joke he heard" will only produce a picture of several people who appear to be talking, whereas adding "country view" to the text prompt generates an image with explicit yet magnificent scenery in the background. Users can change the artist's name and background information based on their preferences.

\begin{figure}[h]
    \centering
    \includegraphics[width = 0.8\linewidth]{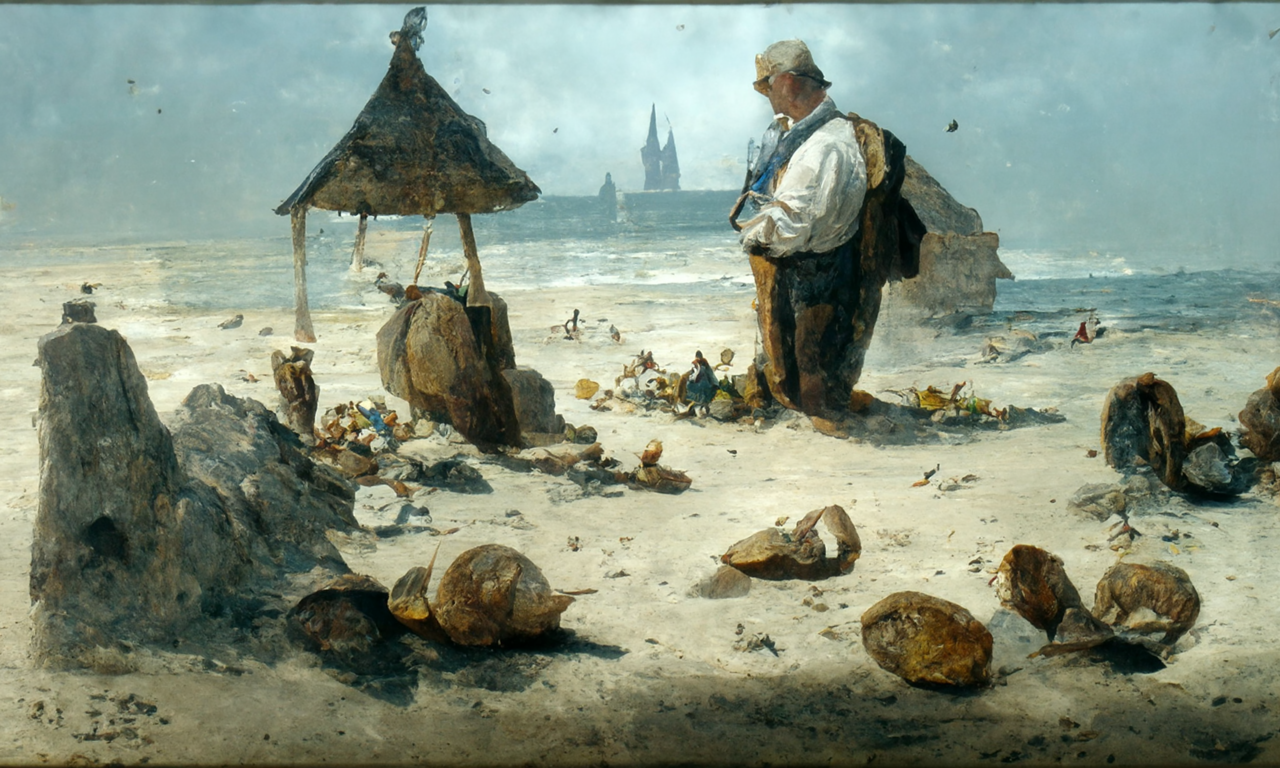}
    \caption{Prompt ``a boy is picking up shells on a beach," generated using Disco Diffusion}
    \label{shellpic}
\end{figure}

Once Disco Diffusion is able to generate images with reasonably appropriate information based on the text prompt, we use it as a source of common sense knowledge to suggest additional keywords for the sentence that follows. For instance, consider the prompt, "a boy is picking up shells on a beach." This generates an image of a boy with a bag picking up shells on a beach with an ocean backdrop, as shown in Figure~\ref{shellpic}. While the image contains requested elements such as "boy," "shells," and "beach," it also introduces new but highly relevant elements such as "bag" and "ocean." This is similar to how when people read stories, they frequently fill in details based on their imagination. The extra elements from the generated images are used to add an "imagination" component to the story generation pipeline. Object detection algorithms are used to detect potential keywords from generated images, which are then presented to users as suggestions for creating the next part of the story.

\subsubsection{Implementation}
There are about 115 parameters in Disco Diffusion. We mainly focus on settings that directly impact our output images.

{\bf Clip$\_$guidance$\_$scale:} This variable tells the model how strongly CLIP will follow the prompt. We set it to 5,000 as recommended.

{\bf Steps:} Each step (or iteration) involves the model looking at a subset of images called ``cuts" and calculating the ``direction" in which the images should be guided so that the generated results are relevant to the prompt. We set this to 250.

{\bf n$\_$batches:} We generate three images for each prompt. Users can choose which one is ideal for that story.
  
\begin{table}[h]
    \centering
    \begin{tabular}{l|c|l}
    \textbf{Item}       &\textbf{Number}  &\textbf{\begin{tabular}[c]{@{}l@{}}Confidence \\ level\end{tabular}}\\
    {$\rhd$ }horse               &5                 & \mybar{0.729}\\
    {$\rhd$ }bird                &2                  & \mybar{0.719}\\
    {$\rhd$ }person              &42                 & \mybar{0.694}\\
    {$\rhd$ }handbag             &2                  & \mybar{0.656}\\
    {$\rhd$ }\dots{}             &\dots{}            &\dots{}
    \end{tabular}
    \caption{
         The table includes items detected based on images generated through Disco Diffusion with their corresponding number detected and confidence level. }
    \label{odTable}
\end{table}

\begin{figure*}[h]
    \centering
    \includegraphics[width = \linewidth]{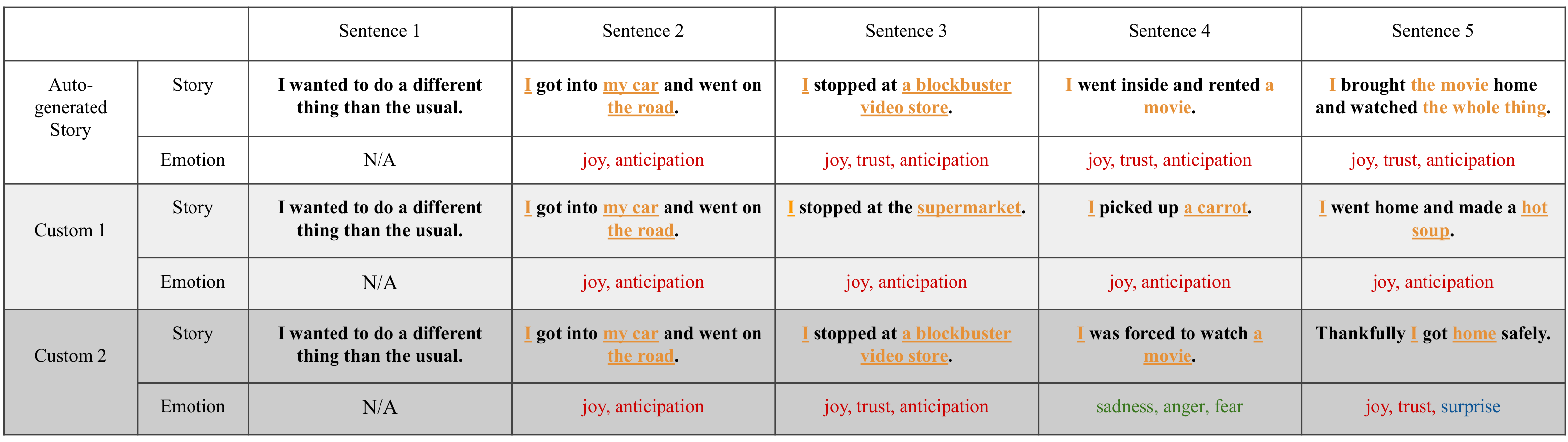}
    \caption{Several generated stories based on the same background sentence. The underlined words in orange indicate the keywords entered}
    \label{tableExaple}
\end{figure*}
\subsection{Key Object Detection}
\begin{figure*}[h]
    \centering
    \includegraphics[width = 0.9\linewidth]{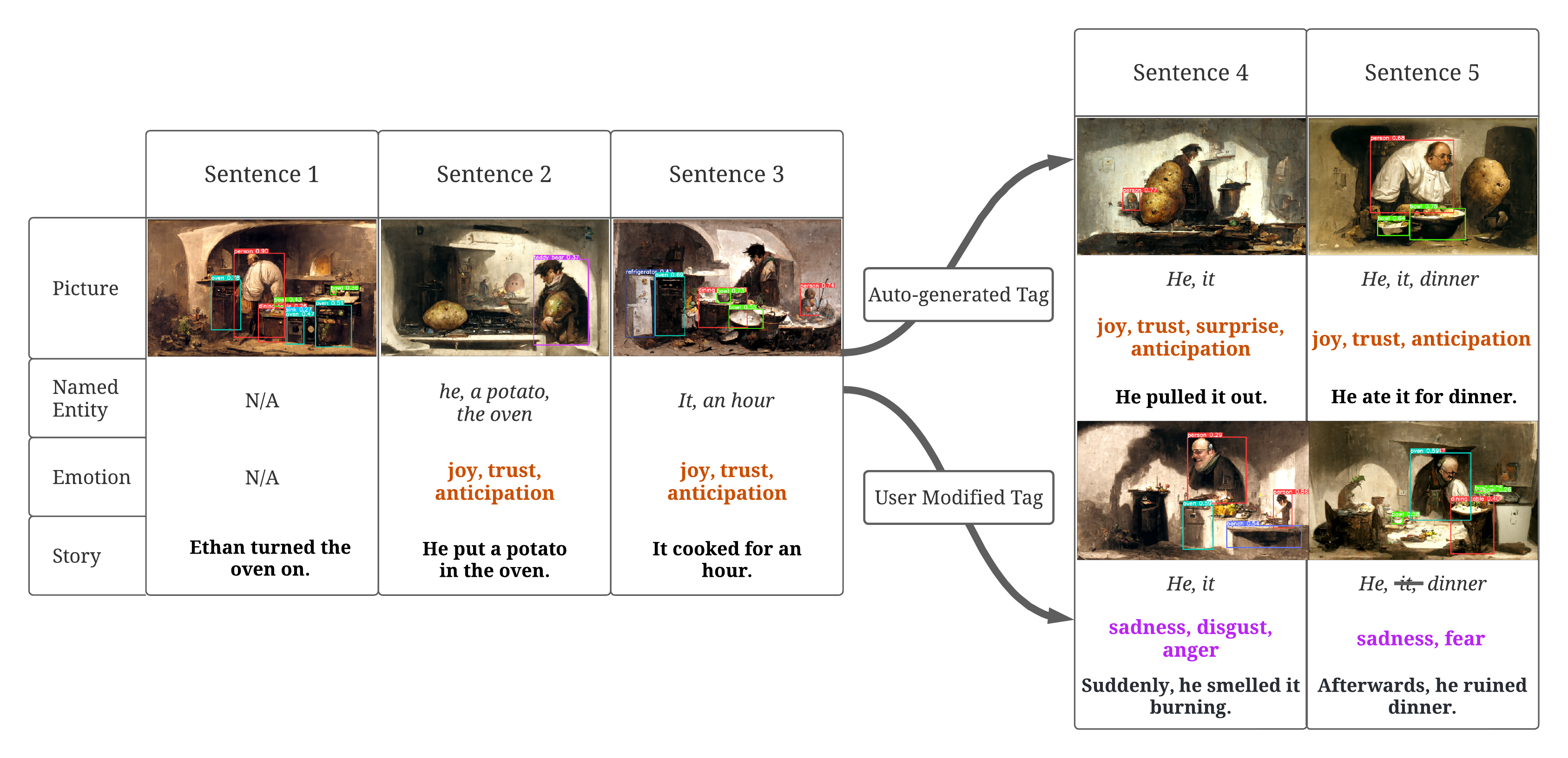}
    \caption{Story fragments generated by the proposed method with two prompt configurations.}
    \label{Example-overview}
\end{figure*}

We explored using YOLOv5 and Faster R-CNN~\cite{https://doi.org/10.48550/arxiv.1506.01497} for object detection. In general, YOLOv5 outperforms Faster R-CNN in inference speed due to its smaller model size. On the other hand, faster R-CNN with ResNet-50 FPN~\cite{https://doi.org/10.48550/arxiv.1612.03144} backbone can provide better performance in terms of mAP (i.e., mean average precision) improvement. Both models are pre-trained with the COCO dataset~\cite{https://doi.org/10.48550/arxiv.1405.0312}. Although these models are only trained on real-world images, they perform impressively on the paintings generated by Disco Diffusion. We suspect this is due to the fact that, as long as the artistic style isn't too abstract, the objects in the generated images share characteristics with real-world objects. 

With the same prompt, Disco Diffusion can generate a batch of images. We process all images created for object detection and summarize a corresponding dictionary of the detected object with their confidence levels. As a result, the user will have a better sense of potential objects associated with the current story and will be more likely to include them in the text prompt. 

\subsubsection{Implementation}
We use YOLOv5x because it produces the best mAP results among the YOLOv5 family. The COCO dataset is used to pre-train YOLOv5 and Faster R-CNN, resulting in approximately 80 detectable objects. To improve the reliability of the results, we set the Faster R-CNN threshold to 0.4. The final result is sorted using the confidence score IOU (Intersection over Union).
\begin{figure}[h]
    \centering
    \includegraphics[width = \linewidth]{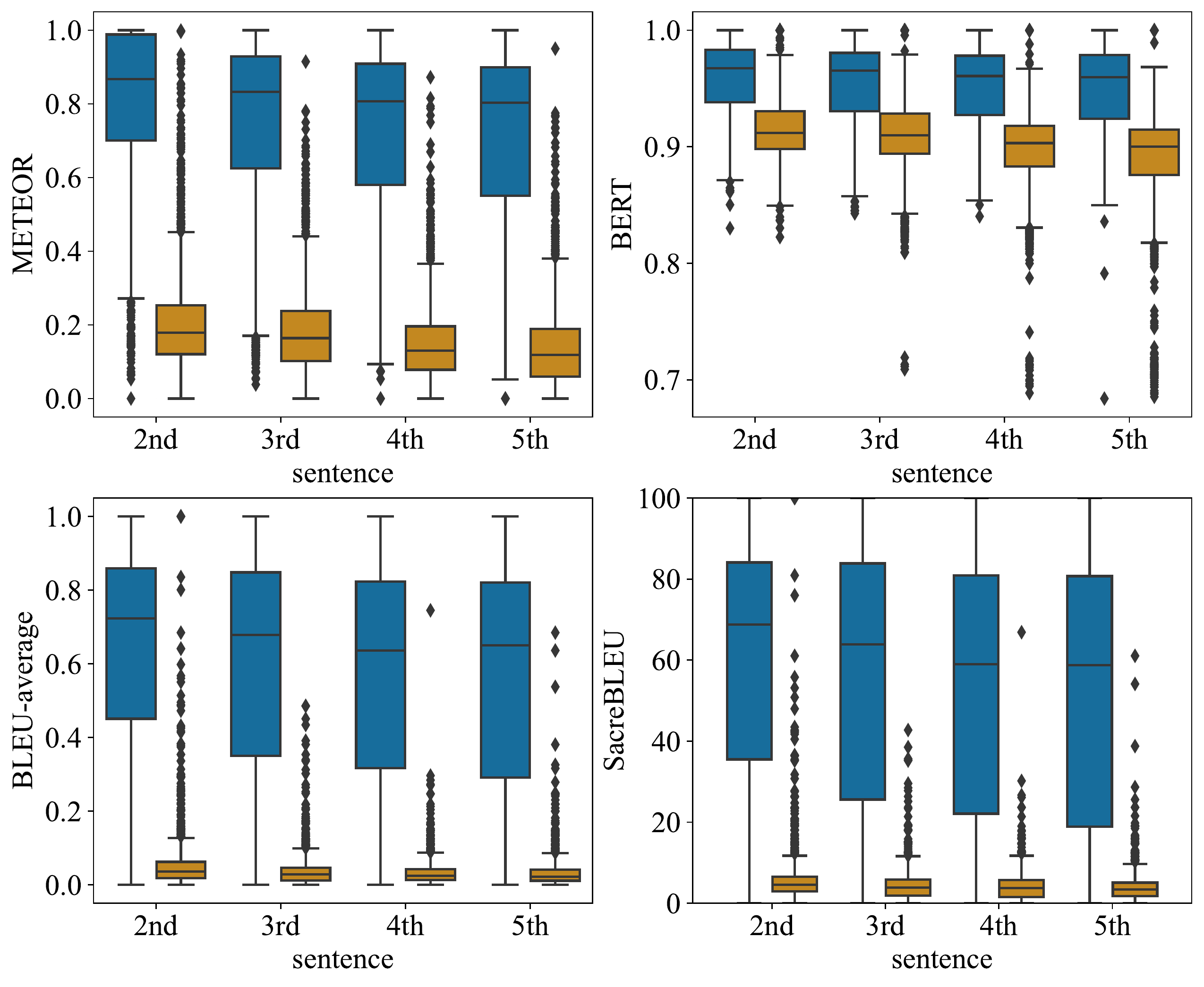}
    \caption{Performance distribution of the baseline model and the prompt-optimized model in 3,748 sets of experiments under different metrics The blue box on the left side of each figure represents our method, and the orange on the right side represents the baseline model.}
    \label{metricCompare-baseline}
\end{figure}

\section{Example Outputs}
Figure~\ref{tableExaple} demonstrates a story generation example using various keywords and emotions using our pipeline. The pipeline can generate a reasonable story based on the provided information by modifying the input prompts of the last two sentences (i.e., the twist and ending parts of the story) in both examples. This demonstrates both the capability of our pipeline and its sensitivity to emotion and keywords. In addition, we demonstrate a full visual storytelling report in Figure~\ref{Example-overview} by combining generated images provided by the image generation pipeline.

\section{Evaluation and Discussion}
To assess the efficacy of a controlled story generation framework based on categorical emotion and keywords, we compared our approach to a baseline model fine-tuned solely based on the existing story. We trained another BERT-based model, hereinafter referred to as BERT$_{\textsc{Cur Emo}}$, to generate sensible emotion keywords as input. This model is a fine-tuned emotion predictor based on the 11,129 emotion-sentence pairs in the Story Commonsense dataset. This model, unlike the BERT$_{\textsc{Next Emo}}$ mentioned in the Categorical Emotion Inference section, aims to detect emotion information for a given (current) sentence. The macro AUC-ROC score reached 0.78 during training, leading us to believe that  BERT$_ {\textsc{Cur Emo}}$ is adequate for identifying and extracting emotions from sentences. To ensure that the models can apply prior knowledge to the inference process of story generation, both our pipeline and the baseline will use the same T5$_{\textsc{BASE Fine-tuned CommonGen}}$ and adapt the inputs to the prompt format used in the corresponding fine-tuning phase.

In order to obtain the validity of the prompts we added to the input, we used the average of BLEU~\cite{lin-och-2004-orange} scores with n-grams ranging from 1 to 4 and BERT-scores~\cite{https://doi.org/10.48550/arxiv.1904.09675} as basic metrics in our evaluation. In addition, we use METEOR~\cite{banarjee2005} and SacreBLEU\cite{post-2018-call} as supplements to compensate for the lack of stemming and synonym matching, as well as standard exact word matching, when comparing the generated content with the ground truth.

We present the experimental results in Figure~\ref{metricCompare-baseline}  and find that, compared to the baseline model, which relies solely on story context for inference, our pipeline shows consistent improvements in all metrics under the same model by introducing emotions and keywords as prompts.

Story generation methods based on emotion and keywords can present statistically higher upper bounds, regardless of how the metrics are implemented or what is tested. As shown in Figure~\ref{metricCompare-baseline}, for BLEU-average scores, our approach improves the story generation task by an average of 62\%. It also reduces the number of samples with zero BERT scores significantly. The average increase in BERT scores is also 6.7\%, and the rate of score fluctuation is reduced. This implies that our approach has the potential to improve many aspects of story generation significantly.

\section{Future Work}
We are interested in making several improvements in the future. We hope to find a better prompt format to fully utilize the potential of language models for the generation task. We hope that, with the assistance of Chain of Thought Prompting~\cite{https://doi.org/10.48550/arxiv.2201.11903}, the model will be able to generate a story with more sophisticated reasoning and causal relationship. The resulting story will then feel more natural and logical. Another area that can be improved is the emotion classifier. At the moment, the model for predicting emotions in the following sentence can only achieve around 60\% accuracy for each of Plutchik's emotion categories which can be further improved. 

The image pipeline has the potential to improve as well. Disco Diffusion, in particular, has difficulty producing images of creatures when configured incorrectly. While we can generate reasonable images by incorporating artists who frequently include corresponding objects and creatures in their paintings as a result of CLIP's recognition capabilities, the Diffusion model will produce nonsense if we do not provide a "reference" artist to paint. Furthermore, because they are only pre-trained with the COCO dataset and real-world objects, YOLOv5 and Faster R-CNN can detect a limited number of objects. In the future, we can use YOLOv5 or Faster R-CNN to implement few-shot learning to increase the number of detected objects~\cite{jimaging8020018}.

\section{Conclusion}
We present a novel pipeline for creating visual stories based on Plutchik's Wheel of Emotions model of emotional knowledge and keyword information. This method assists visual narrative designers in creating and iterating stories in a controlled manner. Our method also includes a complete narrative text-to-image generation pipeline. Experiments show that stories generated by using keywords as prompts have a higher level of grammatical and logical consistency, as well as being more consistent with human writing habits. This work, we believe, is an important step toward a more comprehensive and functional language model-based storytelling tool.

\bibliography{anthology}
\bibliographystyle{aaai}
\end{document}